\documentclass[11pt]{article}

\usepackage[paper=a4paper,margin=1in]{geometry}
\usepackage{latexsym, graphicx}
\usepackage{amsmath, amssymb, mathtools, amsthm, bm, longtable, array, tabularx, subcaption}
\newtheorem{theorem}{Theorem}
\newtheorem{definition}{Definition}

\newtheorem{corollary}{Corollary}
\newtheorem{proposition}{Proposition}

\usepackage{appendix}
\usepackage{verbatim}
\usepackage{cite}
\usepackage[colorlinks,linkcolor={blue!80!black},citecolor={blue!50!black},urlcolor={red!80!black}]{hyperref}
\usepackage{placeins}
\usepackage{pgfplots}
\usepackage{booktabs}
\pgfplotsset{compat=1.17}

\usepackage{xcolor}
\definecolor{softgrey}{rgb}{0.5, 0.5, 0.5}
\definecolor{darkgreen}{rgb}{0.02, 0.53, 0}

\usepackage[ruled]{algorithm2e}

\usepackage{xcolor}

\newcommand{\fref}[1]{Figure~\ref{#1}}

\newcommand{\ID}{\mathbb{D}}

\newcommand{\IH}{\mathbb{H}}

\newcommand{\IR}{\mathbb{R}}
\newcommand{\IS}{S}

\newcommand{\IZ}{\mathbb{Z}}

\newcommand\citep{\cite}
\newcommand\be{\begin{equation}}
\newcommand\ee{\end{equation}}
\newcommand\bea{\begin{eqnarray}}
\newcommand\eea{\end{eqnarray}}

\renewcommand\comment[1]{}

\newcommand\tr{\mathrm{tr}}
\renewcommand\ker{\mathrm{ker}}
\newcommand\im{\mathrm{im}}

\newcommand{\varstr}[2]{\vrule height #1 depth #2 width0pt}

\begin{document}
\thispagestyle{empty}
\renewcommand{\thefootnote}{\fnsymbol{footnote}}

\begin{titlepage}

\begin{center}

\textbf{
{\LARGE Learning to be Simple}
}

\vspace{0.3in}

\textbf{\large
Yang-Hui He$^{\,a,b}$\footnote{\texttt{hey@maths.ox.ac.uk}},
Vishnu Jejjala$^{\,c}$\footnote{\texttt{v.jejjala@wits.ac.za}},
Challenger Mishra$^{\,d}$\footnote{\texttt{cm2099@cam.ac.uk}},
Em Sharnoff$^{\,e,f}$\footnote{\texttt{github@max.sharnoff.org}}
}

\vspace{0.2in}

${}^a$ \textit{London Institute for Mathematical Sciences, Royal Institution, London W1S 4BS, UK}\\
${}^b$ \textit{Merton College, University of Oxford, OX1 4JD, UK}\\
\vskip0.25cm

${}^c$  \textit{Mandelstam Institute for Theoretical Physics, School of Physics, NITheCS, and CoE-MaSS,
University of the Witwatersrand, Johannesburg, South Africa}
\vskip0.25cm

${}^d$  \textit{Department of Computer Science \& Technology, University of Cambridge, CB3 0FD, UK}
\vskip0.25cm

${}^e$ \textit{Department of Computer Science, University of Oxford, OX1 3QG, UK}\\
${}^f$  \textit{Christ Church, University of Oxford, OX1 1DP, UK}
\vskip0.25cm

\end{center}

\vspace{0.1in}

\begin{abstract}
\noindent
In this work we employ machine learning to understand structured mathematical data involving finite groups and derive a theorem about necessary properties of generators of finite simple groups.
We create a database of all $2$-generated subgroups of the symmetric group on $n$-objects and conduct a classification of finite simple groups among them using shallow feed-forward neural networks.
We show that this neural network classifier can decipher the property of simplicity with varying accuracies depending on the features.
Our neural network model leads to a natural conjecture concerning the generators of a finite simple group.
We subsequently prove this conjecture. This new toy theorem comments on the necessary properties of generators of finite simple groups. We show this explicitly for a class of sporadic groups for which the result holds. Our work further makes the case for a machine motivated study of algebraic structures in pure mathematics and highlights the possibility of generating new conjectures and theorems in mathematics with the aid of machine learning.

\end{abstract}

\end{titlepage}

\tableofcontents

\renewcommand{\thefootnote}{\arabic{footnote}}
\setcounter{footnote}{0}

\section{Introduction: Machine learning and symmetries}\label{intro}
Machine learning is an increasingly ubiquitous tool for studying a wide range of problems from self-driving cars and drug design to many electron systems in quantum chemistry and protein folding \textit{in vivo}.
However, thus far machine learning has played a  smaller r\^ole in developing pure mathematics.
Since the injection of machine learning into investigations of algebraic geometry in the context of theoretical physics
\cite{He:2017aed,Carifio:2017bov,Krefl:2017yox,Ruehle:2017mzq,He:2017set,He:2018jtw,Ruehle:2020jrk,jain2022physics}, there has been an explosion of activity to machine learn various aspects of the topology and geometry of Calabi--Yau manifolds~\cite{Bull:2018uow,Bull:2019cij,Berglund:2021ztg,Ashmore:2019wzb,peifer2020learning, Anderson:2020hux, Douglas:2020hpv, Jejjala:2020wcc, Larfors:2021pbb, Ashmore:2021ohf, Larfors:2022nep,Berglund:2022gvm}, algebra~\cite{He:2019nzx,Bao:2020nbi,amoros2021machine,Dechant:2022ccf}, knot theory~\cite{hughes2020neural,Jejjala:2019kio,Gukov:2020qaj,Craven:2020bdz,Craven:2021ckk,Craven:2022cxe}, combinatorics~\cite{He:2020fdg,Bao:2021ofk}, and number theory~\cite{Alessandretti:2019jbs,He:2020eva,He:2020kzg}, etc.
The present work has a two-fold purpose.
We investigate the following questions:
(i) can one learn the structure of mathematics and let artificial intelligence help the intuition of a mathematician along~\cite{He:2021oav,davies2021advancing}?
(ii) how does one develop machine learning architectures that can identify structures in mathematical datasets which are difficult to observe with the human eye?
The aim is to develop new machine driven methodologies and architectures that can study such \textit{synthetic} data, as opposed to \textit{real world} data. 

One fundamental algebraic structure to examine through the lens of machine learning is a group.
Groups provide a mathematical description of the symmetries of a system and are a guiding principle in our descriptions of Nature.
Noether's theorems~\cite{Noether:1918zz} establish that conserved quantities arise from the symmetries of a theory.
For example, the conservation of energy is the consequence of the translational invariance of a system in time, the conservation of momentum is the consequence of the translational invariance of a system in space, and the conservation of angular momentum is the consequence of the invariance of a system under spatial rotations.
Similarly, conserved currents in electromagnetism originate from the $U(1)$ gauge symmetry of quantum electrodynamics.
The particles in the Standard Model are organized according to how they transform, namely in representations of certain Lie groups associated to the gauge symmetries.
Identifying the underlying symmetries of a system and assessing their meaning is of paramount importance in understanding the physics~\cite{arnol2013mathematical,weinberg1995quantum}.

The analogous argument can be made for many machine learning endeavours.
Knowledge of the symmetries of a dataset, real world or otherwise, is central to identifying correlations within the data.
Indeed, making machine learning inferences using datasets is made more tractable by incorporating known invariances of the dataset into the models \textit{ab initio}.
This could, for instance, be accomplished by embedding the invariances into the architecture of a neural network~\cite{cohen2016group, kondor2018generalization}, or in the kernel of a Gaussian process.
This way, various seemingly disconnected parts of the parameter space at play, albeit connected by these invariances, inform each other.
This has the practical advantage of reducing computational costs, improving generalisation, and has led to many real world applications.
The abiding principle is that a cat is a cat regardless of how it is viewed and this equivalence should be in built where possible.
These applications exploit the relationship between model invariances and the dynamics of optimisation, leading to improved generalisation.

Preliminary studies of machine learning the algebraic structure of groups and rings were initiated for finite groups~\cite{He:2019nzx,He:2021oav} and for Lie groups~\cite{Chen:2020jjw}.
For specific algebraic structures, the reader is also referred to~\cite{Bao:2020nbi,He:2020fdg,amoros2021machine,peifer2020learning,Dechant:2022ccf,Cheung:2022itk,lee2023datascientific}.
One key motivation of our present work is to advance these investigations to a deeper level.
With a view towards building a machine driven detector of algebraic structures (and groups in particular), we ask if \textit{interpretable} neural networks can study different properties of a group.
Such properties can range from the order of a group (or group element), to more involved computations such as the invariant ring of a group.
In this work, we concern ourselves with finite simple groups.

In Section~\ref{sec1}, we review some general results on finite simple groups, and describe the subclass of groups we study in this paper, \textit{i.e.}, two generated subgroups of the symmetric group $S_n$ with examples.
We also explain motivations behind our various representations for these groups, and their limitations.
In Section~\ref{sec:results}, we present our machine learning outcomes as well as a proposition that we were able to extract from our the machine learning investigations.
Finally, we conclude with the discussion in Section~\ref{discussion}.
Appendices~\ref{sec:inputs} and~\ref{sec:datasets} describe details of the datasets of groups we employ for our machine learning endeavours, while Appendix~\ref{sec:arch} presents the neural network architectures.

\section{Machine learning simplicity}\label{sec1}

Recall that a group is \textit{simple} if it does not admit any non-trivial normal subgroups.
Following decades of effort, the finite simple groups are completely classified: they are cyclic groups of prime order $\IZ_p$, or alternating groups $A_n$ with $n>4$, or belong to one of $16$ infinite families of groups of Lie type (plus the related Tits group), or are one of $26$ exceptional cases called sporadic groups, the largest of which is the Fischer--Griess Monster, the source of Moonshine~\cite{gorenstein2007finite}.
Indeed, the classification of the finite simple groups initially relied on computational technology to establish the existence and uniqueness of certain sporadic groups.

In~\cite{He:2019nzx}, relatively shallow neural networks as well as support vector machines were used to distinguish simple groups from non-simple one to $99\%$ accuracy without the AI knowing anything about the usual techniques; this beckoned the question as to whether there is underlying new mathematics and constituted a motivation for the present study.
In particular, the Cayley multiplication tables of all finite groups up to size $70$ (there are $602$) were taken.
Next, random permutations were performed on each (since Cayley tables are only defined up to permutations) such that more permutations are included for the simple groups (since there are many more finite groups that are non-simple).
This created a \textit{balanced} database of $60\,000$ examples of $70 \times 70$ matrices (a group of size $n$ would have all entries in $\{1, 2, \ldots, n\}$ and all tables are padded with $0$ where necessary)
labelled as ``simple'' or ``non-simple'', with $50\%$ each.
Remarkably, when flattened and represented as points in $\IR^{70^2}$, a support vector machine with Gaussian kernel was able to \textit{separate} them to $99\%$ accuracy.
This led to a proto-conjecture that in the space of finite groups, the simple and non-simple groups are thus separated and can be so classified.

One shortcoming of using the Cayley table is that these grow as the square of the group order and working in $\IR^{n^2}$ limits the computational power.
In the paper, we focus on building a multi-layer perceptron model that can classify whether finite groups are simple using a much more succinct representation.
This gives a two-fold advantage: (i) it will allow the exploration of more groups; and (ii) it will cross-check whether the simple/non-simple separation is truly underlying some deep mathematics and not just an artifact of Cayley tables.

Now, in the literature there are some age old results regarding the simplicity of groups.
Burnside's theorem, Sylow's test, and detecting zeroes from the character table are notable examples.
Such classical results are helpful in assessing a small sample of possible finite simple groups.
One na\"{\i}ve algorithm that would determine simplicity of a group would list out all possible non-trivial subgroups of a given group and sequentially check if any of those are normal.
State of the art deterministic algorithms that test the simplicity of a finite group are computationally non-trivial even though they might incorporate known classical theorems about finite simple groups.\footnote{ 
One method is to compute the character table and spotting the positions of $0$s. We grant that these are polynomial complexity~\cite{BERNSTEIN2004727}, but our main motivation is to uncover new structures in simple groups using AI rather than to find faster algorithms in their detection.
}

Non-deterministic algorithms are faster at establishing simplicity.
In this paper, we provide an example of a non-deterministic multi-layer perceptron classifier that works in polynomial time in the size of the inputs.\\

\noindent\textbf{Conjecture (Dixon 1969)~\cite{dixon1969probability}}: Two randomly chosen elements of a finite simple group $G$ generate $G$ with
probability $\rightarrow$ 1 as $|G|\rightarrow\infty$.\\

Noting that every finite group is a subgroup of the symmetric group $S_n$ for some $n$ and that every simple group is $2$-generated~\cite{aschbacher1984some},  Dixon's conjecture motivates our study of $2$-generated subgroups of $S_n$, and consider criterion about their simplicity~\cite{dixon1969probability,shalev2001asymptotic}. In the experiments, groups were randomly generated by two permutations drawn at random from $S_n$. We use these as generators of a finite group and query a neural network classifier with the question of simplicity of the resulting group. We conduct a number of experiments with different representations. In one set of experiments, we use the full permutation representation of both the generators (which are required to be unequal to each other and identity). In the second set of experiments, we only use traces and determinants of generators (as they are representation invariant quantities). In a final experiment we only use the orders of the group elements and the order of the group as features. This is motivated by Theorem~\ref{theorem1}, which says that finite simple groups can be characterised by these integers. Further details about the experiments are in   Appendix~\ref{appendix}. 

\section{Learning outcomes and a conjecture}\label{sec:results}
\begin{figure}[ht!]
\centering
\begin{tikzpicture}
\begin{axis}[
    xlabel={Percentage of dataset},
    ylabel={Validation accuracy},
    xmin=0,   xmax=100,
    ymin=0.5, ymax=1.0,
    xtick={0,20,40,60,80,100},
    ytick={0.5,0.6,0.7,0.8,0.9,1.0},
    legend pos=south east,
]
\addplot[color=darkgreen, mark=square] coordinates {
    (4.6,.7533)
};
\addlegendentry{$n = 8$}

\addplot[color=blue, mark=square] coordinates {
    (2,.5163)(5,.8254)(10,.8235)(20,.828)(35,.9032)(50,.972)(80,.9512)(100,.96)
};
\addlegendentry{$n = 7$}

\addplot[color=magenta, mark=square] coordinates {
    (5,.521)(10,.527)(20,.539)(35,.6364)(50,.865)(80,.8654)(100,.96)
};
\addlegendentry{$n = 6$}

\addplot[color=orange, mark=square] coordinates {
    (5,.5923)(10,.5526)(20,.5624)(35,.6359)(50,.7239)(80,.8689)(100,.8891)
};
\addlegendentry{$n = 5$}

\end{axis}
\end{tikzpicture}
\caption{\textsf{
    Validation accuracy for varied amount of datasets for each $n$. Percentages are of the total dataset, after balancing but before splitting to allow $k$-fold cross validation.}
}
\label{fig:accuracies}
\end{figure}

\subsection{Machine learning experiments}\label{exptssec}
\subsection*{Experiment 1}
Our first experiment takes, as input, a pair of matrices as elements of the symmetric group $S_n$.
We mark whether the group generated by the pair as simple or non-simple accordingly.
All results described here were generated via the cross-validation process described in Appendix~\ref{sec:training-datasets}. \autoref{fig:accuracies} shows the individual average validation accuracies at the end of training for each portion of dataset for $n$. Only a single data point is available for $n = 8$ (at $4.6\%$), due to computational restrictions from such a large dataset.

In general, there seem to be two different patterns in the validation accuracies above: for $n = 5$ and $6$, the validation accuracy appears to increase in a roughly linear fashion with the percent of the dataset given, ending between $85\%$ to $95\%$ accuracy. For $n = 7$, however --- and this effect is still visible with $n = 8$ to some extent --- the validation accuracy jumps up to $82\%$ at only $5\%$ of the full dataset. There is a slight decrease as the percent of dataset increases for $n = 7$ from $60\%$ to $80\%$; this is likely due to the variability in the final accuracy for cross-validation runs at $n = 7$. We will discuss this further shortly.

Putting these patterns aside, the models appear to become highly accurate when given greater portions of the dataset --- finishing at $89\%$, $96\%$, and $96\%$ validation accuracy for $n = 5$, $6$, and $7$ respectively. Excluding an outlier in the final result for $n = 7$ gives an average of $99\%$ validation accuracy on the full dataset. \autoref{fig:training-curves} displays the models' loss and accuracy on the training set during training, for each cross-validation run combined. The curves for $n = 5$ fit with the typically expected images; the other two sets merit further comments.

\begin{figure}[ht!]
\centering
\input{training-curves.tex}
\caption{\textsf{Loss and accuracy on training data while training on full datasets, for $n \in \{ 5, 6, 7 \}$. Bold curves give the average value at that epoch across all cross-validation runs.}}

\label{fig:training-curves}
\end{figure}

A particular effect discovered in the training runs for $n = 6$ is that the model appears to take some time to ``get off the ground" --- \textit{i.e.}, some amount of training with slow progress is required before learning can accelerate. This effect was equally visible in the validation loss and accuracy. Multiple initializers were tried as replacements in an attempt to mitigate this effect, but it did not improve. Once training accelerated, the models appeared to behave normally.

Because epochs count the number of times the entire dataset has been used for training, there were also different requirements for each $n$ in the number of epochs run for. For example, $n = 7$ only required six epochs, which partially explains why the per-epoch variability is more visible. Also present in $n = 7$ were occasional regressions during training, where one of the cross-validation runs would suddenly increase in loss and decrease in both training and validation accuracy. This consistently affected one or two cross-validation runs, entirely at random; even though the individual datasets used for each cross-validation run were kept consistent across repeated experimentation, the particular ``failing'' runs were not consistent. In the particular case shown in \autoref{fig:training-curves}, the ``failing'' run finished with a heavy bias for negative results (\textit{i.e.}, indicating that a group is not simple). The class accuracies for that trial were $98\%$ and $4\%$, respectively.
We were unable to identify a cause, though it happens to be simply to manually identify when training a model, given knowledge of the previous behavior. There were insufficient data to determine whether this effect would occur at $n = 8$.

\begin{figure}[ht!]
\centering
\begin{tikzpicture}[scale=0.8]
\begin{axis}[
    title={$n = 5$},
    name=n5,
    xlabel={Epoch},
    ylabel={Absolute Prediction Error (\%)},
    xmin=0, xmax=30,
    ymin=0, ymax=15,
    xtick={0,5,10,15,20,25,30},
    ytick={0,2.5,5,7.5,10,12.5,15},
    legend pos=north east,
    legend columns=2,
    legend style={nodes={scale=0.8, transform shape}},
]

\addplot[color=darkgreen] coordinates {
    (5,3.074)(10,2.291)(15,2.231)(20,2.495)(25,3.166)(30,2.971)
};
\addplot[color=darkgreen,dashed,forget plot] coordinates {
    (5,11.246)(10,12.160)(15,6.796)(20,5.655)(25,5.766)(30,5.031)
};
\addlegendentry{100\%}

\addplot[color=blue] coordinates {
    (5,4.577)(10,1.993)(15,1.783)(20,1.951)(25,2.562)(30,2.959)
};
\addplot[color=blue,dashed,forget plot] coordinates {
    (5,5.888)(10,13.174)(15,12.055)(20,8.598)(25,8.737)(30,8.907)
};
\addlegendentry{80\%}

\addplot[color=magenta] coordinates {
    (5,3.281)(10,1.827)(15,1.958)(20,1.878)(25,1.067)(30,0.912)
};
\addplot[color=magenta,dashed,forget plot] coordinates {
    (5,6.197)(10,9.109)(15,10.413)(20,9.711)(25,10.827)(30,11.215)
};
\addlegendentry{50\%}

\addplot[color=orange] coordinates {
    (5,3.518)(10,2.031)(15,2.477)(20,2.532)(25,2.257)(30,2.464)
};
\addplot[color=orange,dashed,forget plot] coordinates {
    (5,7.288)(10,5.908)(15,4.280)(20,6.718)(25,6.612)(30,7.146)
};
\addlegendentry{35\%}

\end{axis}

\begin{axis}[
    title={$n = 6$},
    at={(n5.outer north east)},anchor=outer north west,
    xlabel={Epoch},
    ylabel={Absolute Prediction Error (\%)},
    xmin=0, xmax=30,
    ymin=0, ymax=15,
    xtick={0,5,10,15,20,25,30},
    ytick={0,2.5,5,7.5,10,12.5,15},
    legend pos=north east,
    legend columns=2,
    legend style={nodes={scale=0.8, transform shape}},
]

\addplot[color=darkgreen] coordinates {
    (5,2.652)(10,2.983)(15,4.596)(20,4.704)(25,4.753)(30,4.811)
};
\addplot[color=darkgreen,dashed,forget plot] coordinates {
    (5,3.953)(10,7.276)(15,12.156)(20,5.370)(25,5.219)(30,5.168)
};
\addlegendentry{100\%}

\addplot[color=blue] coordinates {
    (5,2.983)(10,2.598)(15,1.304)(20,3.621)(25,4.380)(30,4.696)
};
\addplot[color=blue,dashed,forget plot] coordinates {
    (5,3.250)(10,3.649)(15,11.804)(20,8.616)(25,5.407)(30,5.354)
};
\addlegendentry{80\%}

\addplot[color=magenta] coordinates {
    (5,5.472)(10,2.604)(15,2.865)(20,2.851)(25,1.989)(30,1.175)
};
\addplot[color=magenta,dashed,forget plot] coordinates {
    (5,2.009)(10,3.540)(15,3.526)(20,4.608)(25,6.997)(30,11.121)
};
\addlegendentry{50\%}

\addplot[color=orange] coordinates {
    (5,2.416)(10,3.318)(15,2.487)(20,3.072)(25,2.808)(30,1.985)
};
\addplot[color=orange,dashed,forget plot] coordinates {
    (5,2.419)(10,2.874)(15,3.404)(20,3.174)(25,2.973)(30,3.751)
};
\addlegendentry{35\%}

\end{axis}   
\end{tikzpicture}
\caption{\textsf{Error in predictions of accuracy from simplicity (dashed) versus parity (solid) during training for $n = 5$ (left) and $n = 6$ (right).}}
\label{fig:simple-vs-parity-predictions}
\end{figure}

In order to gain further insights into understanding simplicity of finite groups, we now describe further machine learning experiments we conducted using alternate features. 

\subsection*{Experiment 2}
\begin{figure}[t!]
\centering
  \includegraphics[width=0.3\textwidth]{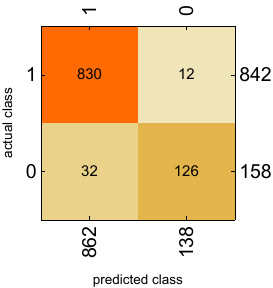}\hspace{20pt} \includegraphics[width=0.3\textwidth]{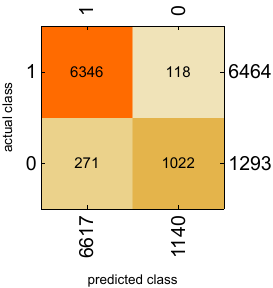}\\
  \includegraphics[width=0.3\textwidth]{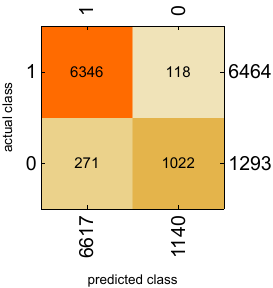}\hspace{20pt} \includegraphics[width=0.3\textwidth]{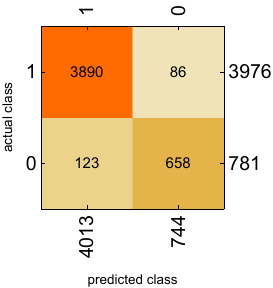}\\
  (L) Training performance; (R) Test performance
\caption{A classifier predicts simplicity of a group based on orders and traces of group generators and the property of being Abelian. The plot shows a confusion plot where the class $0$ stands for simple groups and $1$ for nonsimple groups. Using a neural network model and  $1000$ training points, class accuracies were found to be $\approx 97\% $ and $ 85\%$, respectively, on the training set (left) and $\approx 97\%$ and $86\%$ on the test set (right). We use $1000$ (top row) and $4000$ (bottom row) points for training. }
\label{fig:expt2}
\end{figure}

In this experiment we choose the features to be the traces and determinants of the two generators and the binary property of the group being Abelian.  We used a multi-layer perceptron model with three hidden layers with $1000$, $500$, and $200$ nodes respectively, and logistic sigmoid activation. We used an ADAM optimiser. The dataset of size $\approx 8500$  was split into a training set of sizes size $1000$, $4000$. The remaining data was used for validation in both cases.  The results are described in~\fref{fig:expt2}. 

\begin{figure}[t!]
\centering
 \includegraphics[width=0.3\textwidth]{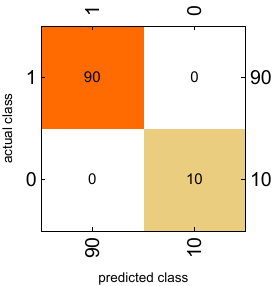}\hspace{20pt} \includegraphics[width=0.3\textwidth]{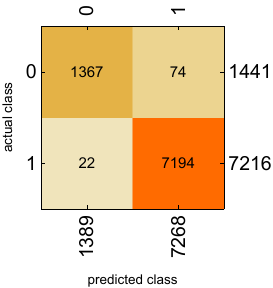}\\
  \includegraphics[width=0.3\textwidth]{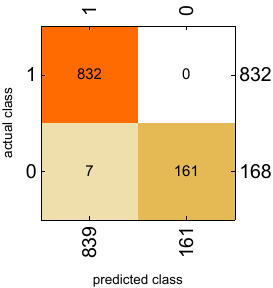}\hspace{20pt} \includegraphics[width=0.3\textwidth]{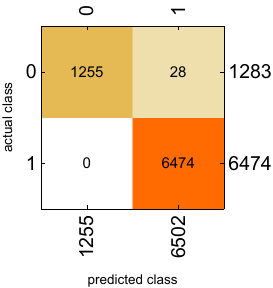}\\
  (L) Training performance; (R) Test performance
\caption{A classifier predicts simplicity of a group based on orders of group elements and order of the group. The plot shows a confusion plot where the class $0$ stands for simple groups and $1$ for nonsimple groups. Using a neural network model when using only $100$ training points (top) and $1000$ training points (bottom), class accuracies on the test set were found to be $\approx 99\%$.}
\label{expt3}
\end{figure}
\subsection*{Experiment 3}
In this experiment we chose the features to be the orders of the group elements and the order of the group. We used a multi-layer perceptron model with three hidden layers with $1000$, $500$, and $200$ nodes respectively, and logistic sigmoid activation. We used an ADAM optimiser. The dataset of size $\sim8500$  was split into a training set of size $1000$, and a test set of $\sim7500$. We repeated the experiment using a smaller architecture with 10 times fewer nodes and a training set of size $100$. The results are described in~\fref{expt3}. 

The second and third experiments were particularly illuminating in terms of choice of features. The new features in these experiments were the traces and determinants of the generators as opposed to the full generators; and some combination of two other features, the group order and the property of being Abelian. 

With these outcomes, it is interesting to consider a mathematical statement that might capture these experiments. We therefore consider necessary conditions on traces and determinants of generators of finite simple groups; hinted by our machine learning outcomes above. This manner of investigation has yielded different conjectures using an alternative approach~\cite{mishra2023mathematical}.

\subsection{A machine guided mathematical conjecture}\label{math-res}
In Experiments 1--3 of Section~\ref{exptssec} we have built simple feed-forward multi-layer perceptron models to predict the simplicity of a $2$-generated group given merely the generators, or their properties. Our learning outcomes were modest with predictive accuracies $\sim 80\%$. 
In a traditional approach, one can conceive of a straightforward methodology for resolving the question of the simplicity of the resulting group. This could be seen to involve two key steps, the first of which is to freely generate a group from the two generators. Following this, one could list all non-trivial subgroups of this group, and check for normal subgroups. Both these key steps are computationally intensive. Therefore, it is impressive that a simple feed-forward multi-layer perceptron model should address this question to any reasonable degree of accuracy. We now ask the following question: can similar learning outcomes be attained using alternate representations for the generators? These could involve high level properties of the generators given a representation --- traces and determinants are examples of such properties. If a similar or better learning outcome is obtained using such the above properties as alternate features, this is perhaps indicative of an underlying mathematical relationship between such generator properties and the property of simplicity. This indeed turns out to be the case: \autoref{fig:expt2} and \autoref{expt3} show learning outcomes using different features. 
In this section, we present such a mathematical relationship as a conjecture motivated by the above learning outcome, for which we then provide a proof.
This AI-guided mathematical conjecture formulation is very much in the spirit of~\cite{He:2018jtw,davies2021advancing,mishra2023mathematical}.

First, we recall a theorem for characterising finite simple groups~\cite{vasil2009characterization}. Let us define the set of orders of group elements as 
\begin{equation}
\pi(G):= \{\mbox{Ord}(g) :  g\in G\} \ .   
\end{equation}

\begin{theorem}\label{theorem1}
A finite simple group $G$ is completely characterised by the order of group elements $\pi(G)$ and the order (size) of the group $G$. 
\end{theorem}

Theorem~\ref{theorem1} implies that for $G$ and $H$ simple finite groups, $G\simeq H$ iff $|G|=|H|$ and $\pi(G)=\pi(H)$ (as sets). This is indicative that perhaps determinants of the generator could play a crucial role in discovering a new mathematical relationship between the generators and the property of simplicity. In fact, this also provides an alternate representation for our classification problem. Computing $\pi(G)$ for a group $2$-generated $G$ requires further computation. When our ML models are trained using these alternate features, it is not surprising that the learning outcomes are nearly perfect. \fref{expt3} shows the learning outcome. However, this is at additional computational cost, since generating such representations involves computing $\pi(G)$ and computing $|G|$ requires. 

Furthermore, we make an observation from studying our dataset. Before doing this, we introduce a preliminary notion~\cite{burness2017simple}. Let $G$ be a subgroup of a symmetric group acting as permutation of a set $\Omega$, as is with the case with all groups we consider here.
We recall the standard definitions that for an element $\alpha \in \Omega$, the stabilizer $G_\alpha$ of $\alpha$ is the set of all group elements that fix $\alpha$.
On the contrary, the set of all fixed points for a group element $x \in G$ is denoted $C_\Omega(x)$.
In short,
\begin{align}
\nonumber
    G_\alpha &:=
    \{x \in G : x(\alpha) = \alpha\} \ , \\
    C_\Omega(x) &:=
    \{\alpha \in \Omega :
    x(\alpha) = \alpha \} \ .
\end{align}
Then,
\begin{definition}
    The fixed point ratio of 
    $x \in G$, denoted by $\text{fpr}(x)$ (which is of course implicitly dependent upon $\Omega$), is the proportion of points in 
    $\Omega$ fixed by $x$, \textit{i.e.},
    \[
    \text{fpr}(x) 
    = \frac{|C_\Omega(x)|}{|\Omega|} \ .
    \]
\end{definition}

For our dataset of all the $2$-generated subgroups of $S_n$ up to $n \leq 10$, 
we notice that whenever a resulting group is simple, the number of fixed points of either of its two generators is never $n-2$ or $n-4$ for $n\ge5$. 
Now, all our group elements are $n \times n$ permutation matrices (in particular matrices with only $0$ and $1$) acting on $\{1,2,\ldots,n\}$. Hence, the number of fixed points is counted by the number of 1s on the diagonal.
Hence, this observation hints towards the existence of a conjecture regarding traces (or equivalently, number of fixed points) and signs of permutations. 

\begin{proposition}\label{conjecture}
(conjectured from machine learning)
Consider $2$-generated subgroups ${\IH}\subset {\IS}_n$ ($n\ge 5$) with distinct non-trivial generators in the permutation representation.
That is, consider 
$\sigma_{i=1,2} (\ne e) \in \IS_n$ as $n \times n$ permutation matrices. 
If ${\IH}$ is a simple group, then
\begin{itemize}\label{thm}
    \item[1.] $\det(\sigma_i)= 1$, and,
    \item[2.] $\tr(\sigma_i) \in \{1, 2, \ldots, n\}\setminus \{n-4,n-2,n-1,n\}$. 
\end{itemize}
In particular, if $\tr(\sigma_i)= n-4 $, then
${\IH}=\mathbb{D}_{2n}\subseteq \mathbb{A}_n \vartriangleleft \mathbb{S}_n,n\ge5$.
That is $\IH$ is the dihedral group, and thus not simple.\\
\end{proposition}

We can prove this conjecture which is inspired from our machine learning experiments.
\begin{proof}
First, since $\IH$ is a subgroup of $S_n$ as represented by permutation matrices, the determinant of all group elements is equal to $\pm 1$.
Suppose, without loss of generality, $\det(\sigma_1)=-1$.
It is straightforward to check that the following defines a homomorphism from ${\IH}$ to $(\IZ/(2\IZ),+\text{mod}\, 2)$.
\begin{center}
    $\psi : {\IH}\longrightarrow \{0,\ 1\}$\\[5pt]
$\psi(\sigma)=\begin{cases}
        ~~ 0, \text{~if~} \det(\sigma)=1 ~, \\
        ~~ 1, \text{~if~} \det(\sigma)=-1 ~.
\end{cases}$\\[10pt]
\end{center}
By the first isomorphism theorem, ${\IH}/\ker(\psi)\simeq \im(\psi)$, with $\ker(\psi)\vartriangleleft {\IH}$. Clearly, $\psi(e)=0$, and $\psi(\sigma_1)=1$ (by assumption). Therefore, $\im(\psi)=\{0,\ 1\}=\mathbb{Z}/(2\mathbb{Z})$. As such, $\ker(\psi)$ is an index two subgroup of ${\IH}$, and therefore normal. Hence, ${\IH}$ is not simple. 
Thus the determinant of both $\sigma_1$ should by 1.
This proves the first part of the proposition.

To prove the second part, let's assume $\mathbb{H}$ is simple. We begin by observing that $\tr(\sigma_i)\ne n$ since $e$ is the only element that has trace $n$ and $\sigma_i\ne e$ by assumption. 
Progressing further, $\tr(\sigma_i)\ne n-1$ as that would require a single $0$ and 1s everywhere else on a diagonal, which is a singular matrix.
Finally, we in fact have that  $\tr(\sigma_i)\ne n-2$ since then $\det(\sigma_i)=-1$, coming from the ${\scriptsize \left( \begin{matrix} 0 & 1 \\ 1 & 0 \end{matrix} \right)}$ block.

Now consider the case of $\tr(\sigma_i)= n-4$ and  $\det(\sigma_i)= 1$. We will show that $\sigma_i^2=e$. If this were to be the case, $\IH$ would be a group generated by involutions. This would imply that ${\IH}$ is the dihedral group of order $2m$, $\ID_{2m}$ (where for for $m=1,2$ they are, $\IZ/(2\IZ)$ and $\IZ/(2\IZ) \times \IZ/(2\IZ)$). 
In any case, $\IH$ would not simple (except for the trivial case of $m=1$, which we excluded by having $n \geq 5$). 

To finish the proof,  it remains show that 
\[
P(n):=((\det(\sigma)=1)\wedge (\tr(\sigma)=n-4)\implies \sigma_i=\sigma^{T})
\]
holds. We do this by induction.
We can enumerate to see that $P(4)$ holds. Now assume $P(n)$ holds. Let $\widehat{\sigma}\in\mathbb{A}_{n+1}$, so that $\det(\widehat{\sigma})=1$ (by definition of the alternating group), and $\tr(\widehat{\sigma})=(n+1)-4$. We must show that $P(n+1)$ holds by showing $\widehat{\sigma}=\widehat{\sigma}^{T}$. Note that each $\widehat{\sigma}$ can be obtained from a ${\sigma}$ with the introduction of a $1$ in the diagonal at one of $(n+1)$ possible positions (reflecting the fact that $\left| \mathbb{A}_{n+1}\right|=(n+1)\left| \mathbb{A}_{n}\right|$). 
Note that this doesn't alter the determinant but increases the trace by $1$. 
It therefore follows that $\widehat{\sigma}=\widehat{\sigma}^T$ since ${\sigma}={\sigma}^T$, establishing that $P(n+1)$ holds, completing the proof. 
\end{proof}

The following corollary is a restatement of the above written in terms of fixed point ratios.

\begin{corollary}\label{corollary}
Let $\IH$ be a finite simple group with the generating set containing only two distinct group elements in a permutation representation of degree $n$. Then, the fixed point ratio ($\mathrm{fpr}$) of any such generator cannot equal $2^i/n,\ \forall i \in\{0,1,2\}$.
\end{corollary}

\begin{table}[!t]
    \centering
    \begin{tabular}{llcc}
        \toprule
        Sporadic group & generators & trace & $\{n-2^k\}_{k=0}^{2}$ \\
        \midrule
        M$_9$ & (1,4,9,8)(2,5,3,6), \\
              & (1,6,5,2)(3,7,9,8) & 1, 1 & 5, 7, 8 \\
        M$_{10}$ & ( 1, 9, 6, 7, 5)( 2,10, 3, 8, 4), \\
                &  ( 1,10, 7, 8)( 2, 9, 4, 6) & 0, 2 & 6,8,9 \\
        M$_{11}$ & ( 1, 2, 3, 4, 5, 6, 7, 8, 9,10,11), \\
                & (3,7,11,8)(4,10,5,6) & 0, 3 & 7,9,10 \\
        M$_{12}$ & (1,2,3,4,5,6,7,8,9,10,11), \\
                & (3,7,11,8)(4,10,5,6), \\
                & (1,12)(2,11)(3,6)(4,8)(5,9)(7,10) & 1, 4, 0 & 8,10,11 \\
        M$_{21}$ & ( 1, 4, 5, 9, 3)( 2, 8,10, 7, 6)\\
        &(12,15,16,20,14)(13,19,21,18,17), \\
        &( 1,21, 5,12,20)( 2,16, 3, 4,17)\\
        &( 6,18, 7,19,15)( 8,13, 9,14,11)
         & 0, 3 & 17,19,20 \\
        M$_{22}$ & ( 1, 2, 3, 4, 5, 6, 7, 8, 9,10,11) \\
        & (12,13,14,15,16,17,18,19,20,21,22), \\
                & ( 1, 4, 5, 9, 3)( 2, 8,10, 7, 6)\\
                &(12,15,16,20,14)(13,19,21,18,17),\\
                & ( 1,21)( 2,10, 8, 6)( 3,13, 4,17)\\
                &( 5,19, 9,18)(11,22)(12,14,16,20)& 1, 1, 1 & 18,20,21 \\
        M$_{23}$ & ( 1, 2, 3, 4, 5, 6, 7, 8, 9,10,11,12,\\
        & 13,14,15,16,17,18,19,20,21,22,23), \\
                & ( 3,17,10, 7, 9)( 4,13,14,19, 5)\\
                & ( 8,18,11,12,23)(15,20,22,21,16) & 1, 1 & 19, 21,22 \\
        M$_{24}$ & ( 1, 2, 3, 4, 5, 6, 7, 8, 9,10,11,12,13,\\
        & 14,15,16,17,18,19,20,21,22,23), \\
                &  ( 3,17,10, 7, 9)( 4,13,14,19, 5)\\
                & ( 8,18,11,12,23)(15,20,22,21,16), \\
                & ( 1,24)( 2,23)( 3,12)( 4,16)\\
                &( 5,18)( 6,10)( 7,20)(8,14)\\
                & ( 9,21)(11,17)(13,22)(15,19) & 1, 4, 0 & 20, 22, 23 \\
    \end{tabular}
    \caption{\textsf{Properties of Mathieu Group generators. All the generators are of positive signature and the number of fixed points is never $n-2^k,~k\in\{0,1,2\}$, ala proposition~\ref{thm}.}}
    \label{tab:sporadic_table}
\end{table}
\newpage
\section{Discussion}\label{discussion}
In this work, we demonstrate that standard off the shelf machine learning tools such as neural networks can help determine the simplicity of a group with relatively high accuracies and produce novel insights. We conduct a number of machine learning experiments with different mathematical features to determine simplicity. When using permutation representations of group generators, we find modest learning outcomes. In another experiment, using element orders as features alongside the order of the group, we demonstrate remarkable class accuracies of close to $99\%$ in~\fref{expt3}, reflective of a known result for finite simple groups, \textit{viz.}, Theorem~\ref{theorem1}. 
The success our experiments gives further confidence that the ability of machine-learning (in particular support vector machines) to distinguish simple/non-simple groups when trained on Cayley tables \cite{He:2019nzx} was not a mere artifact of data representation, but truly underlies interesting mathematics.

From these experiments, we distill Proposition~\ref{conjecture}, which we then prove using properties of finite simple groups. The result restricts choices of $2$-generating sets for finite simple groups. We recast this result in terms of fixed point ratios of group generators in Corollary~\ref{corollary}. Bounds on fixed point ratios are an important topic of consideration. The above observation feeds directly into studies of fixed point ratios of groups, which have been studied extensively over many decades~\cite{burness2017simple}. The result places restrictions on the traces of generators of any finite simple group. The computational advantage this observation yields is  $\approx10^{-4}\%$ when looking for a generator of the Monster group in the smallest faithful irreducible representation which occurs at $n=196\,883$. The more interesting point is that the determinant and trace of the generators conspire in some cases to form a dihedral group. 
Corollary~\ref{corollary} holds for all finite simple groups in the permutation representation. As such it holds for sporadic groups in their permutation representations. In Table~\ref{tab:sporadic_table}, we check consistency of our results for the Mathieu group. For this, we note that the traces and determinants of known generators of the Mathieu group are consistent with the allowed values in Corollary~\ref{corollary}.
We can as well study other presentations of finite groups, and we leave this to future work.

The prospect of finding novel mathematical results is tantalising in the age of AI. There are a number of emerging pathways for mathematical research in light of machine learning. In this work, we exploit one such pathway which involves learning a mathematical property (in this case simplicity) using standard machine learning architectures, and exploiting insights from the learning process to distill a theorem. Often, tools from machine interpretability such as relevance scores can aid in this process resulting in new mathematical insights or theorems~\cite{He:2018jtw,Craven:2020bdz,He:2021oav,davies2021advancing}. Complementary approaches have been proposed in the recent literature, relying on an organisational principle for mathematical statements, resulting in novel conjectures which could often be proved using domain expertise~\cite{mishra2023mathematical}. 
\vspace{-10pt}

\section*{Acknowledgments}\label{ack}
YHH would like to thank STFC for grant ST/J00037X/2, the Leverhulme Trust for a project grant, as well as Joseph Chuang and Radha Kessar for many helpful discussions.
VJ is supported by the South African Research Chairs Initiative of the Department of Science and Innovation and the National Research Foundation.
CM is supported by the Accelerate Programme for Scientific Discovery, at the Computer Laboratory, University of Cambridge. CM would like to thank Dami\'an Kaloni Mayorga Pen\~a, Aditya Ravuri, Subhayan Roy Moulik for helpful discussions. 
The authors would like to thank the Isaac Newton Institute for Mathematical Sciences for support and hospitality during the program ``Black holes: bridges between number theory and holographic quantum information'' when work on this paper was undertaken; this work was supported by EPSRC grant number EP/R014604/1.

\vspace{-20pt}

\appendix

%
\begin{table}[]
\centering
\begin{tabular}{| c || c | c | c |}
\hline\varstr{16pt}{10pt} Group Id & Name & Simple? & Generators \\ \hline
\varstr{16pt}{10pt} [ 1, 1 ] & 1 & No & $ \left[ \begin{array}{cccc} 1 & 0 & 0 & 0 \\ 0 & 1 & 0 & 0 \\ 0 & 0 & 1 & 0 \\ 0 & 0 & 0 & 1 \\ \end{array}\right] \left[ \begin{array}{cccc} 1 & 0 & 0 & 0 \\ 0 & 1 & 0 & 0 \\ 0 & 0 & 1 & 0 \\ 0 & 0 & 0 & 1 \\ \end{array}\right] $\\ \hline
\varstr{16pt}{10pt} [ 2, 1 ] & $\IZ_2$ & Yes & $ \left[ \begin{array}{cccc} 1 & 0 & 0 & 0 \\ 0 & 1 & 0 & 0 \\ 0 & 0 & 1 & 0 \\ 0 & 0 & 0 & 1 \\ \end{array}\right] \left[ \begin{array}{cccc} 0 & 1 & 0 & 0 \\ 1 & 0 & 0 & 0 \\ 0 & 0 & 1 & 0 \\ 0 & 0 & 0 & 1 \\ \end{array}\right] $\\ \hline
\varstr{16pt}{10pt} [ 3, 1 ] & $\IZ_3$ & Yes & $ \left[ \begin{array}{cccc} 1 & 0 & 0 & 0 \\ 0 & 1 & 0 & 0 \\ 0 & 0 & 1 & 0 \\ 0 & 0 & 0 & 1 \\ \end{array}\right] \left[ \begin{array}{cccc} 0 & 0 & 1 & 0 \\ 1 & 0 & 0 & 0 \\ 0 & 1 & 0 & 0 \\ 0 & 0 & 0 & 1 \\ \end{array}\right] $\\ \hline
\varstr{16pt}{10pt} [ 4, 1 ] & $\IZ_4$ & No & $ \left[ \begin{array}{cccc} 1 & 0 & 0 & 0 \\ 0 & 1 & 0 & 0 \\ 0 & 0 & 1 & 0 \\ 0 & 0 & 0 & 1 \\ \end{array}\right] \left[ \begin{array}{cccc} 0 & 0 & 0 & 1 \\ 1 & 0 & 0 & 0 \\ 0 & 1 & 0 & 0 \\ 0 & 0 & 1 & 0 \\ \end{array}\right] $\\ \hline
\varstr{16pt}{10pt} [ 6, 1 ] & $S_3$ & No & $ \left[ \begin{array}{cccc} 0 & 1 & 0 & 0 \\ 1 & 0 & 0 & 0 \\ 0 & 0 & 1 & 0 \\ 0 & 0 & 0 & 1 \\ \end{array}\right] \left[ \begin{array}{cccc} 0 & 0 & 1 & 0 \\ 1 & 0 & 0 & 0 \\ 0 & 1 & 0 & 0 \\ 0 & 0 & 0 & 1 \\ \end{array}\right] $\\ \hline
\varstr{16pt}{10pt} [ 24, 12 ] & $S_4$ & No & $ \left[ \begin{array}{cccc} 0 & 1 & 0 & 0 \\ 1 & 0 & 0 & 0 \\ 0 & 0 & 1 & 0 \\ 0 & 0 & 0 & 1 \\ \end{array}\right] \left[ \begin{array}{cccc} 0 & 0 & 0 & 1 \\ 1 & 0 & 0 & 0 \\ 0 & 1 & 0 & 0 \\ 0 & 0 & 1 & 0 \\ \end{array}\right] $\\ \hline
\varstr{16pt}{10pt} [ 8, 3 ] & $\ID_8$ & No & $ \left[ \begin{array}{cccc} 0 & 1 & 0 & 0 \\ 1 & 0 & 0 & 0 \\ 0 & 0 & 1 & 0 \\ 0 & 0 & 0 & 1 \\ \end{array}\right] \left[ \begin{array}{cccc} 0 & 0 & 1 & 0 \\ 0 & 0 & 0 & 1 \\ 1 & 0 & 0 & 0 \\ 0 & 1 & 0 & 0 \\ \end{array}\right] $\\ \hline
\varstr{16pt}{10pt} [ 4, 2 ] & $\IZ_2 \times \IZ_2$ & No & $ \left[ \begin{array}{cccc} 0 & 1 & 0 & 0 \\ 1 & 0 & 0 & 0 \\ 0 & 0 & 1 & 0 \\ 0 & 0 & 0 & 1 \\ \end{array}\right] \left[ \begin{array}{cccc} 1 & 0 & 0 & 0 \\ 0 & 1 & 0 & 0 \\ 0 & 0 & 0 & 1 \\ 0 & 0 & 1 & 0 \\ \end{array}\right] $\\ \hline
\varstr{16pt}{10pt} [ 12, 3 ] & $A_4$ & No & $ \left[ \begin{array}{cccc} 0 & 0 & 1 & 0 \\ 1 & 0 & 0 & 0 \\ 0 & 1 & 0 & 0 \\ 0 & 0 & 0 & 1 \\ \end{array}\right] \left[ \begin{array}{cccc} 0 & 0 & 0 & 1 \\ 0 & 1 & 0 & 0 \\ 1 & 0 & 0 & 0 \\ 0 & 0 & 1 & 0 \\ \end{array}\right] $\\ \hline
\end{tabular}
\end{table}
\appendix
\section{Machine learning experiments}\label{appendix}

There were therefore many entries in the dataset for groups equivalent under isomorphism, and the number of entries per-group was not changed to be less imbalanced. This was intentional.\footnotemark

\footnotetext{To give a sense of scale, the number of subgroups of $S_n$ for $n = [4,7]$ are $11$, $19$, $56$, $96$. The number of entries generated by our methods for each $S_n$ would be $560$, $14375$, $518\,364$, $25\,401\,551$.}

\subsection{$2$-generated subgroups of $S_n$}\label{sec:inputs}

The input data to our models were the permutation matrices corresponding to the generators of each group. We found this to be the most effective representation of the data, as it allowed for fixed-size inputs and produced better performing models than encoding the generators as permutation vectors.

\subsection*{Example: $2$-generated subgroups of $S_3$}

\begin{table}[h!]
\centering
\begin{tabular}{| c || c | c | c |}
\hline\varstr{16pt}{10pt} Group Id & Name & Simple? & Generators \\ \hline
\varstr{16pt}{10pt} [ 1, 1 ] & 1 & No & $ \left[ \begin{array}{ccc} 1 & 0 & 0 \\ 0 & 1 & 0 \\ 0 & 0 & 1 \\ \end{array}\right] \left[ \begin{array}{ccc} 1 & 0 & 0 \\ 0 & 1 & 0 \\ 0 & 0 & 1 \\ \end{array}\right] $\\ \hline
\varstr{16pt}{10pt} [ 2, 1 ] & $\IZ_2$ & Yes & $ \left[ \begin{array}{ccc} 1 & 0 & 0 \\ 0 & 1 & 0 \\ 0 & 0 & 1 \\ \end{array}\right] \left[ \begin{array}{ccc} 0 & 1 & 0 \\ 1 & 0 & 0 \\ 0 & 0 & 1 \\ \end{array}\right] $\\ \hline
\varstr{16pt}{10pt} [ 3, 1 ] & $\IZ_3$ & Yes & $ \left[ \begin{array}{ccc} 1 & 0 & 0 \\ 0 & 1 & 0 \\ 0 & 0 & 1 \\ \end{array}\right] \left[ \begin{array}{ccc} 0 & 0 & 1 \\ 1 & 0 & 0 \\ 0 & 1 & 0 \\ \end{array}\right] $\\ \hline
\varstr{16pt}{10pt} [ 6, 1 ] & $S_3$ & No & $ \left[ \begin{array}{ccc} 0 & 1 & 0 \\ 1 & 0 & 0 \\ 0 & 0 & 1 \\ \end{array}\right] \left[ \begin{array}{ccc} 0 & 0 & 1 \\ 1 & 0 & 0 \\ 0 & 1 & 0 \\ \end{array}\right] $\\ \hline
\end{tabular}
\caption{\textsf{Subgroups of $S_3$.}}
\end{table}


Model training and validation were performed on subsets of the \textit{balanced} datasets for $5$-fold
cross-validation. The mean final validation accuracy for models trained on different portions of the
balanced dataset for each $n$ is displayed in \autoref{fig:accuracies}, from ``Results''.


\subsection{Datasets}\label{sec:datasets}

The datasets are given by the generating permutation matrices for each subgroup of some $S_n$. Only the two-generated subgroups were selected, but all permutations were included, giving a total $(n!)^2 - n^2$ individual entries in each dataset. The orders of the groups in these datasets were therefore not evenly distributed; the number of each group included in the dataset for $S_4$ and $S_5$ is in \autoref{tab:s4-counts} and \autoref{tab:s5-counts}. It was computationally infeasible to include the same information for $S_6$ and larger.

\begin{table}
    \centering
    \begin{tabular}{| c | c | c | c | c |}
        \hline\varstr{16pt}{10pt} Group Id & Name & Simple? & Count (filtered) & Count (unfiltered) \\ \hline
        \varstr{16pt}{10pt} [ 2, 1 ] & C2 & Yes & 18 & 27\\ \hline
        \varstr{16pt}{10pt} [ 3, 1 ] & C3 & Yes & 24 & 32\\ \hline
        \varstr{16pt}{10pt} [ 4, 1 ] & C4 & No & 30 & 36\\ \hline
        \varstr{16pt}{10pt} [ 4, 2 ] & C2 $\times$ C2 & No & 24 & 24\\ \hline
        \varstr{16pt}{10pt} [ 6, 1 ] & S3 & No & 72 & 72\\ \hline
        \varstr{16pt}{10pt} [ 8, 3 ] & D8 & No & 72 & 72\\ \hline
        \varstr{16pt}{10pt} [ 12, 3 ] & A4 & No & 96 & 96\\ \hline
        \varstr{16pt}{10pt} [ 24, 12 ] & S4 & No & 216 & 216\\ \hline
    \end{tabular}
    \caption{\textsf{Number of each subgroup of $S_4$ that would be present in the dataset.}}
    \label{tab:s4-counts}
\end{table}
\begin{table}
    \centering
    \begin{tabular}{| c | c | c | c | c |}
        \hline\varstr{16pt}{10pt} Group Id & Name & Simple? & Count (filtered) & Count (unfiltered) \\ \hline
        \varstr{16pt}{10pt} [ 2, 1 ] & C2 & Yes & 50 & 75\\ \hline
        \varstr{16pt}{10pt} [ 3, 1 ] & C3 & Yes & 60 & 80\\ \hline
        \varstr{16pt}{10pt} [ 4, 1 ] & C4 & No & 150 & 180\\ \hline
        \varstr{16pt}{10pt} [ 4, 2 ] & C2 x C2 & No & 120 & 120\\ \hline
        \varstr{16pt}{10pt} [ 5, 1 ] & C5 & Yes & 120 & 144\\ \hline
        \varstr{16pt}{10pt} [ 6, 2 ] & C6 & No & 220 & 240\\ \hline
        \varstr{16pt}{10pt} [ 6, 1 ] & S3 & No & 360 & 360\\ \hline
        \varstr{16pt}{10pt} [ 8, 3 ] & D8 & No & 360 & 360\\ \hline
        \varstr{16pt}{10pt} [ 10, 1 ] & D10 & No & 360 & 360\\ \hline
        \varstr{16pt}{10pt} [ 12, 3 ] & A4 & No & 480 & 480\\ \hline
        \varstr{16pt}{10pt} [ 12, 4 ] & D12 & No & 360 & 360\\ \hline
        \varstr{16pt}{10pt} [ 20, 3 ] & C5 : C4 & No & 1440 & 1440\\ \hline
        \varstr{16pt}{10pt} [ 24, 12 ] & S4 & No & 1080 & 1080\\ \hline
        \varstr{16pt}{10pt} [ 60, 5 ] & A5 & Yes & 2280 & 2280\\ \hline
        \varstr{16pt}{10pt} [ 120, 34 ] & S5 & No & 6840 & 6840\\ \hline
    \end{tabular}
    \caption{\textsf{Number of each subgroup of $S_5$ present in the dataset.}}
    \label{tab:s5-counts}
\end{table}

For all datasets, we used the \texttt{SageMath} interface to \texttt{GAP} to analyze the groups generated by each pair of permutation matrices. The categorization of whether a group is simple was paired with the generating matrices to produce the entry for that group.

With most of the symmetric groups examined here, it was feasible to produce the categorization from \texttt{GAP} for all permutation pairs, which was done by simple iteration. Experiments on $S_8$ used a random subset of all possible permutation pairs by taking a random sample of integers $1$ to $(8!)^2$, mapping these values to permutation matrices, and filtering out pairs with the same permutation. The component of this mapping that produces the permutation\footnotemark\ is specified in~\autoref{alg:perm-mapping}.\\

\footnotetext{Constructing the permutation matrix from the corresponding permutation of $n$ elements is trivial; we leave this undiscussed.}

\begin{algorithm}[H]
\SetAlgoLined
\DontPrintSemicolon
\KwIn{An integer $k \in [0,n!)$}
\KwOut{A permutation of the elements ${ 0, 1, .. n }$}
%
%

$\text{vs} \coloneqq [\ ]$ \tcp{The values 0, .. n}
$\text{ps} \coloneqq [\ ]$ \tcp{The final permutation}

\For{$i \leftarrow 0$ \KwTo $n - 1$}{ vs.append($i$) }

\For{$i \leftarrow n$ \KwTo $1$}{
    $ r\coloneqq$ remainder of $\frac{k}{i}$\;
    $k \leftarrow k \mod i$\;
    
    $v \coloneqq \text{vs}[r]$\;
    Remove $\text{vs}[r]$ and shift remaining elements to the left\;
    ps.append($v$)\;
}

\KwRet{ps}

\caption{\textsf{Conversion from $k \in [0,n!)$ to a permutation of $n$ elements.}}
\label{alg:perm-mapping}
\end{algorithm}

A single sample in one of the datasets could be constructed as follows. Suppose we are generating data for the subgroups of $S_4$, with the permutations given by the integer pair $(6, 19)$. The corresponding permutations, as generated by the aforementioned algorithm, are:
\begin{equation}
(2, 1, 0, 3) \text{\ and\ } (3, 1, 2, 0)
\end{equation}

\noindent which generate the following permutation matrices, respectively:
\begin{equation}
\left( \begin{array}{cccc}
0 & 0 & 1 & 0 \\
0 & 1 & 0 & 0 \\
1 & 0 & 0 & 0 \\
0 & 0 & 0 & 1 \\
\end{array} \right)
\text{\ and\ }
\left( \begin{array}{cccc}
0 & 0 & 0 & 1 \\
0 & 1 & 0 & 0 \\
0 & 0 & 1 & 0 \\
1 & 0 & 0 & 0 \\
\end{array} \right)
\end{equation}

\noindent The inputs for the training sample are just the concatenation of each row in the two matrices into a single vector, as shown below:
\begin{equation}
\langle 0,0,1,0,\ 0,1,0,0,\ 1,0,0,0,\ 0,0,0,1,\ 0,0,0,1,\ 0,1,0,0,\ 0,0,1,0,\ 1,0,0,0 \rangle
\end{equation}

\noindent The particular group corresponding to these inputs $S_3$, which is not simple.

\subsection*{Training datasets}
\label{sec:training-datasets}

The generated datasets were unbalanced; only approximately $20\%$ of the entries correspond to simple groups in each dataset.\footnotemark\ From here on, we refer only to the \textit{balanced} datasets, which are chosen by a random selection of the non-simple entries in equal number to the set of simple ones.

\footnotetext{For each value of $n \in \{5,..8\}$, these fractions are $0.1758$, $0.2027$, $0.2315$, and $0.2133$, respectively.}

All experiments were done with $5$-fold cross-validation, with each train/validation split taking from the same balanced subset of the unbalanced dataset. For experimentation on differing amounts of data (\textit{e.g.}, using $50\%$ of the ``full", balanced dataset), all balanced datasets were subsets of the ``full" balanced dataset.

\subsection{Neural network architectures}\label{sec:arch}

\begin{figure}[ht!]
\centering

\begin{tikzpicture}[scale=0.62]
    \tikzstyle{reverseclip}=[insert path={(current page.north east) --
      (current page.south east) --
      (current page.south west) --
      (current page.north west) --
      (current page.north east)}
    ]
    
    \def\layerboxdim{1}
    \def\layerbox#1#2#3{
        \draw (#1,#2) rectangle (#1+\layerboxdim,#2+#3);
    }
    \def\innernode#1#2#3{
        \node at (#1+0.5*\layerboxdim,#2+0.5*\layerboxdim) {\tiny #3}
    }
    \def\dotspath#1#2#3#4{ \path (#1) -- (#2) node[pos=#4,sloped,font=#3] {$\dots$} }
    \def\layer#1#2#3#4#5#6{
        \layerbox{#1}{#2}{#3}
        \draw (#1,#2+\layerboxdim) -- (#1+\layerboxdim,#2+\layerboxdim);

        \dotspath{#1+0.5*\layerboxdim,#2+\layerboxdim}
                 {#1+0.5*\layerboxdim,#2+#3-#4*\layerboxdim}{\large}{0.5};

        \foreach \y [evaluate=\y as \n using int(#6-\y+1)] in {1,2,...,#4} {
            \draw (#1,#2+#3-\y*\layerboxdim) -- (#1+\layerboxdim,#2+#3-\y*\layerboxdim);
            \innernode{#1}{#2+#3-\y*\layerboxdim}{$#5_{\n}$};
        }
        \innernode{#1}{#2}{$#5_1$};
    }

    \def\inx{4}
    \def\hax{\inx+3}
    \def\hbx{\hax+2.5}
    \def\hcx{\hbx+3.5}
    \def\outx{\hcx+3}

    \foreach \y in { 5, 11, 12, 13, 14 }
            \dotspath{\hbx+\layerboxdim,\y+0.5}{\hcx,\y+0.5}{\normalsize}{0.5};
    \dotspath{\hax+1.75,6}{\hax+1.75,11}{\normalsize}{0.5};

    \begin{pgfinterruptboundingbox}
    \path[clip] (\hbx+1.6,5) -- (\hcx-.6,5) -- (\hcx-.6,15) -- (\hbx+1.6,15) -- cycle [reverseclip];
    \path[clip] (\hax+1.25,7.5) -- (\hbx-.25,7.5) -- (\hbx-.25,9.5) -- (\hax+1.25,9.5) -- cycle [reverseclip];
    \end{pgfinterruptboundingbox}

    \layerbox{\inx}{7}{6}
    \foreach \y in {8, 11, 12} \draw (\inx,\y) -- (\inx+\layerboxdim,\y);
    \innernode{\inx}{7}{$x_1$};
    \dotspath{\inx+0.5,8}{\inx+0.5,11}{\Large}{0.5};
    \innernode{\inx}{11}{$x_{\mathbb{N}-1}$};
    \innernode{\inx}{12}{$x_{\mathbb{N}}$};
    \layer{\hax}{5}{10}{4}{h^1}{256}
    \layer{\hbx}{5}{10}{4}{h^2}{256}
    \layer{\hcx}{5}{10}{4}{h^n}{256}
    \layerbox{\outx}{9}{2}
    \draw (\outx,10) -- (\outx+\layerboxdim,10);
    \innernode{\outx}{9}{$\hat{y}_0$};
    \innernode{\outx}{10}{$\hat{y}_1$};

    \foreach \ysa/\xa/\ysb/\xb in {
        {7,11,12}/\inx/{5,11,12,13,14}/\hax, {5,11,12,13,14}/\hax/{5,11,12,13,14}/\hbx,
        {5,11,12,13,14}/\hbx/{5,11,12,13,14}/\hcx,
        {5,11,12,13,14}/\hcx/{9,10}/\outx
    } {
        \foreach \ya in \ysa {
            \foreach \yb in \ysb {
                \draw (\xa+\layerboxdim,\ya+0.5*\layerboxdim) -- (\xb,\yb+0.5*\layerboxdim);
            }
        }
    }

    \draw[thick, decoration={brace,raise=0.15cm}, decorate]
        (\inx,7) -- (\inx,13) node[pos=0.5,xshift=-0.5cm,rotate=90] { $2n^2$ inputs };

    \draw[thick, decoration={brace, mirror, raise=0.15cm}, decorate]
        (\hax,5) -- (\hcx + \layerboxdim,5) node[pos=0.5,yshift=-0.5cm] { $n$ hidden layers };

    \draw[thick, decoration={brace, mirror, raise=0.15cm}, decorate]
        (\outx+\layerboxdim,9) -- (\outx+\layerboxdim,11) node[pos=0.5,xshift=1.2cm] { 2 outputs };

\end{tikzpicture}
\caption{\textsf{Generic architecture of the models used. The generating permutation matrices are flattened and provided as input. Softmax is applied to the two outputs, which give probabilities for whether the group is simple or not.}}
\label{fig:architecture}
\end{figure}

The models used for experimentation were fully-connected neural networks with the two input permutation matrices flattened into a single vector of $2n^2$ boolean values for each $S_n$. Predictions are chosen by the greater of two output values, scaled by the softmax activation function. Hidden layers for all final models consist of $256$ nodes, with all nodes using the ReLU activation function. See \autoref{fig:architecture} above for an illustration.

Optimization was done with stochastic gradient descent (SGD) with Nesterov-accelerated momentum and Mean Squared Error as the loss function. The hyperparameters for SGD (both learning-rate and momentum), the number of hidden layers for each $n$, and the size of the hidden layers were selected by manual hyperparameter search. The results with the best mean validation accuracy across all five validation datasets at the end of training. A hidden layer size of $256$ was chosen for all $n$ as a common best size for $n = 5$ and $n = 6$.

\begin{table}[ht!]
\centering
\begin{tabular}{ |c|c|c|c|c| }
    \hline
    $n$ & Learning rate & Momentum & Num. hidden layers & $\gamma$ \\
    \hline
    5 & 0.05   & 0    & 1 & 0.1 \\
    6 & 0.001  & 0.01 & 3 & 0.05 \\
    7* & 0.01  & 0.1  & 9 & 0.05 \\
    8* & 0.01  & 0.1  & 9 & 0.05 \\
    \hline
\end{tabular}
\caption{\textsf{Hyperparameters used for each dataset.}}
\label{tab:hparams}
\end{table}

For $n = 7$ and $n = 8$, rigorous hyperparameter search was infeasible due to the size of the datasets and the resulting computational restrictions -- less so with the former than the latter. For these datasets, the hyperparameters were initially chosen by extrapolation from those used with the lower two values for $n$, without attempting to produce optimal results. Adjustments were made only until the model was successfully better than random chance. This less rigorous iteration was required as, for some sets of hyperparameters, the models completely fail to perform any meaningful learning (often skewing entirely towards one output or another). This occurred most prominently at $n = 7$, when testing with five or fewer hidden layers. This effect was avoided by increasing the number of hidden layers, which did not prove to be additionally necessary for $n = 8$ (\textit{i.e.}, there was no increase from $n = 7$  to $8$).

In addition to the hyperparameters previously mentioned, learning rate decay was also provided, with the learning rate at each epoch calculated from the previous by:
\begin{equation} \text{lr}_{i+1} = (1 - \gamma) \text{lr}_i \end{equation}
with $\gamma$ as the hyperparameter controlling the rate of decay. The hyperparameters used for each $n$ are given above, in \autoref{tab:hparams}.

\bibliographystyle{JHEP} 
\bibliography{ref}

\end{document}